\def\year{2019}\relax
\documentclass[letterpaper]{article} %DO NOT CHANGE THIS
\usepackage{aaai19}  %Required
\usepackage{times}  %Required
\usepackage{helvet}  %Required
\usepackage{courier}  %Required
\usepackage{url}  %Required
\usepackage{graphicx}  %Required
\usepackage{amsmath}
\usepackage{amsfonts}
\usepackage{multirow}
\usepackage{makecell}

\title{Improving Multilingual Semantic Textual Similarity with Shared Sentence Encoder for Low-resource Languages}
\author{
Xin Tang\textsuperscript{1}\thanks{Corresponding author: Xin Tang (eli.tx@alibaba-inc.com)}, Shanbo Cheng\textsuperscript{1}, Loc Do\textsuperscript{1}, Zhiyu Min\textsuperscript{2}, Feng Ji\textsuperscript{1}, Heng Yu\textsuperscript{1}, Ji Zhang\textsuperscript{1}, Haiqin Chen\textsuperscript{1}  \\
\textsuperscript{1}{Alibaba Group, Hangzhou, China} \\ 
\textsuperscript{2}{Carnegie Mellon University} \\
\{eli.tx, shanbo.csb, dohaloc.mr, zhongxiu.jf, yuheng.yh, zj122146, haiqing.chenhq\}@alibaba-inc.com\\
zhiyum@andrew.cmu.edu
}

\begin{document}
\nocopyright
\maketitle
\begin{abstract}
Measuring the semantic similarity between two sentences (or Semantic Textual Similarity - STS) is fundamental in many NLP applications. Despite the remarkable results in supervised settings with adequate labeling, little attention has been paid to this task in low-resource languages with insufficient labeling. Existing approaches mostly leverage machine translation techniques to translate sentences into rich-resource language.  These approaches either beget language biases, or be impractical in industrial applications where spoken language scenario is more often and rigorous efficiency is required. In this work, we propose a multilingual framework to tackle the STS task in a low-resource language e.g. Spanish, Arabic , Indonesian and Thai, by utilizing the rich annotation data in a rich resource language, e.g. English. Our approach is extended from a basic monolingual STS framework to a shared multilingual encoder pretrained with translation task to incorporate rich-resource language data. By exploiting the nature of a shared multilingual encoder, one sentence can have multiple representations for different target translation language, which are used in an ensemble model to improve similarity evaluation. We demonstrate the superiority of our method over other state of the art approaches on SemEval STS task by its significant improvement on non-MT method, as well as an online industrial product where MT method fails to beat baseline while our approach still has consistently improvements.
\end{abstract}

\section{Introduction}

% add a teaser picture in the upper-right of the first page

Semantic Textual Similarity (STS) is a fundamental task in many Natural Language Processing applications such as question answering, machine translation, semantic search, etc. \cite{cer2017semeval}.  For its importance, there has been a growing interest in developing solutions for the task from both academia and industry. In particular, deep learning techniques have been used extensively in STS under supervised settings \cite{DBLP:journals/tacl/YinSXZ16,DBLP:conf/aaai/PangLGXWC16}. The common approach is to take advantage of pretrained word embeddings such as Word2Vec \cite{DBLP:conf/nips/MikolovSCCD13}
%and GloVe \cite{DBLP:conf/emnlp/PenningtonSM14}
, therefrom a deep neural network is used to extract the sentence representations as well as the interactions between them. Subsequently, a final Multi-Layer Perceptron (MLP) is trained from the representations and interactions to fit the STS label. Despite achieving outstanding performances, this approach requires large amounts of labeling, which restricts its applicability in settings with insufficient labeling, such as low-resource languages like Spanish, Arabic and Thai.

%Although only few works are focusing on the low-resource problem, it is urgent and obligatory in industrial applications. 

%Popular network designs include Recurrent Neural Network (RNN) and Convolutional Neural Network (CNN). RNN processes the sentence sequentially, using the last hidden state as the final sentence representation. On the other hand, CNN captures regional features parallelly with receptive windows. It is already claimed that CNN can achieve comparable results with RNN but features much higher efficiency to obtain sentence representation \cite{gehring2017convolutional}.

Existing approaches to STS in low-resource languages mostly leverage machine translation (MT) techniques. One possible approach is to translate the target sentences to a resource-rich language where a well-trained semantic similarity model can be obtained \cite{tian-EtAl:2017:SemEval}. Even though  this MT based approach provides strong baselines in the past SemEval tasks, it also poses several drawbacks.  First, the translation quality depends highly on the quality of input sentence. SemEval data sets are collected from formal writing sources such as books and newspapers, which can be translated from English to other languages quite accurately. In practice, it is common to observe sentences with informal writing styles including typos, slang or abbreviation. The translation quality of these sentences often degrades~\cite{belinkov2018synthetic}.  Besides possible semantic loss introduced by translation, MT based method clearly abates the efficiency in online services. Another approach \cite{tian-EtAl:2017:SemEval} is to incorporate various language independent features such as sentence length and lexical similarity to achieve an ensemble model which also takes more computing resource in real time.

In this work, we propose a shared encoder framework to perform STS in a target language with insufficient labeling by utilizing annotated data in rich-resource languages. More specifically, we expand a basic monolingual framework for STS to a multilingual one, where an encoder is shared in both languages. In order to alleviate language discrepancy, inspired by machine translation techniques \cite{artetxe2018unsupervised}, we conduct a bi-directional translation task on the shared encoder, together with a shared decoder for both languages. Meanwhile, this translation framework also allows self-translation, which is similar to denoising auto-encoder and can help reserve the original semantics. Due to the shared encoder, one sentence can be encoded into different language semantic spaces by prepending the target language token to the sentence. Finally, a shared Multi-Layer Perceptron (MLP) is trained to fit the pair similarity.

We conduct experiments on an off-line public data set SemEval and industrial data sets from real-world online service. On SemEval data set, we calculate the similarity of Spanish and Arabic pairs and our method consistently beats other non-MT state of art approaches. On the online service in Thai and Indonesian, our method has been deployed and verified in a spoken language data set that its performance prevails MT based one.

Our main contributions are as follows:
\begin{itemize}
    \item We propose a multilingual shared encoder framework to perform STS task in low-resource settings, which effectively leverages annotated data in rich-resource languages, and achieves promising results with little supervision.
    
    \item We employ bi-directional machine translation task to obtain a multilingual encoder. This approach alleviates language discrepancies, as well as captures  general-purpose semantic distributions. 
    
    \item This paper also provide a data augmentation like way by exploiting the shared encoder to obtain sentence representations on different language semantic space and shows its effectiveness on further improving semantic similarity evaluation.
\end{itemize}

\section{Methodology}
% will need to arrange the figures

In this paper, we are focusing on Semantic Textual Similarity (STS) task, where we are given two sentences $s_1=(w_1^1, ..., w_n^1)$ and $s_2=(w_1^2, ..., w_m^2)$ and the goal is to predict their similarity. Our approach includes two stages of training. We firstly learn a shared multilingual sentence encoder through translation task on large parallel corpus. Then, sentence textual semantic task is trained based on the pretrained multilingual encoder with labeled data from both rich-resource language and low-resource language.

%\subsection{Network Architecture}
\subsection{Pretrained Multilingual Encoder}
Our shared multilingual encoder architecture depends on a shared encoder machine translation model. In this section, we describe how we design and train the shared encoder translation model. 

\subsubsection{Shared Encoder Model Architecture}
\begin{figure}[!hbt]
    \centering
	\includegraphics[scale=0.4]{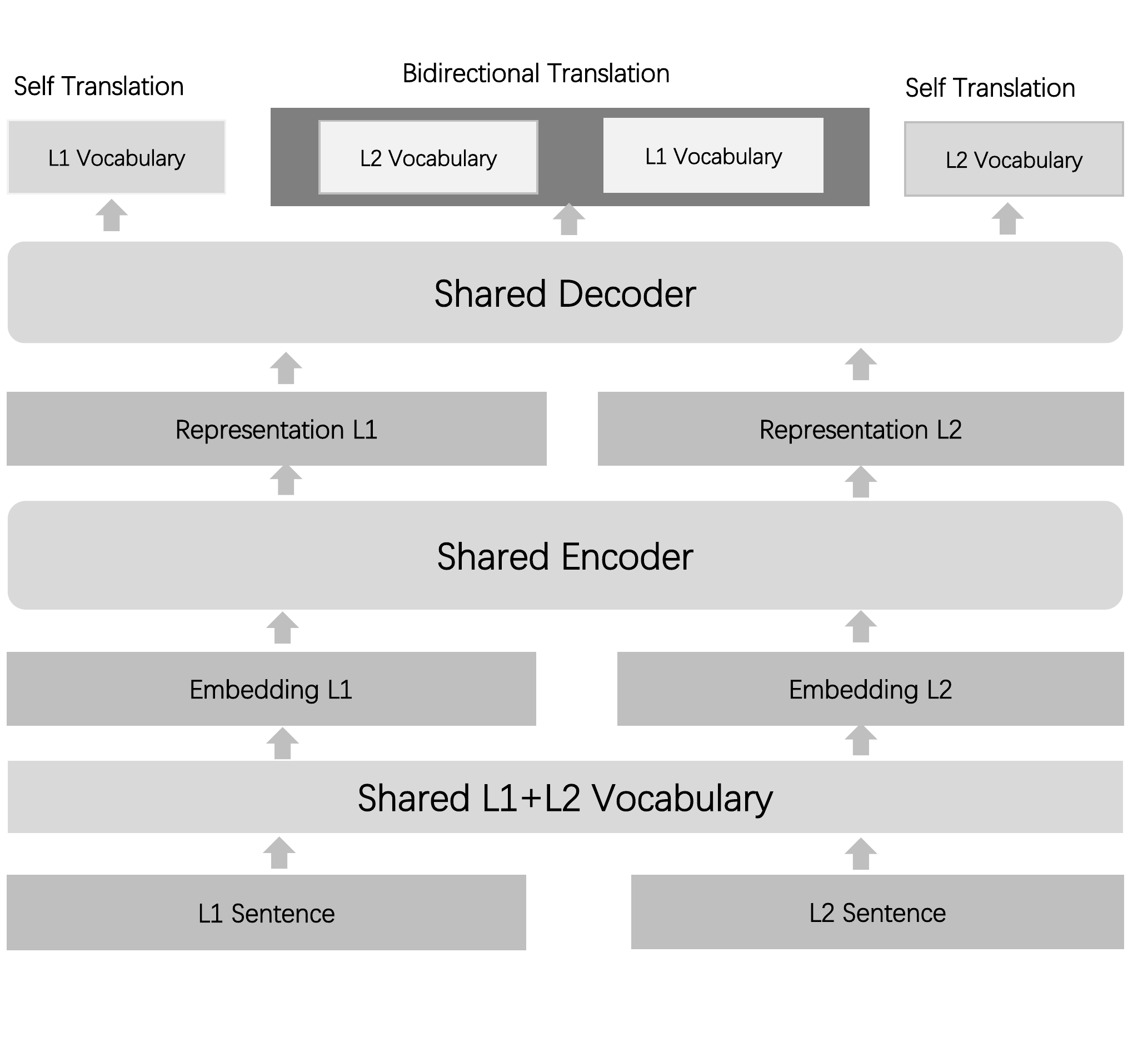}
	\caption{A special case of shared encoder translation model: bidirectional model. Given an input sentence of either language $L_1$ or $L_2$, the model firstly uses a shared vocabulary to get the embedding of each word within the sentence, then the shared encoder will encode the input sentence to get the semantic representation. Lastly, the shared decoder chooses vocabulary $L_1$ or $L_2$ based on the language token in the input sentence, and then generate the final translation. }
	\label{fig:bidrec_translation}
\end{figure}

%\label{sec:sh_enc_model}
We adopt the state-of-the-art transformer architecture \cite{vaswani2017attention} for our translation model. To simplify our explanation, Figure~\ref{fig:bidrec_translation} shows a special example of shared encoder translation model: a bidirectional translation model. Inspired by~\cite{johnson2017google}, the model consists of a shared encoder and a shared decoder. The encoder uses a shared vocabulary for both languages, while the decoder uses two separated vocabularies, one for each language.

Given a sentence to be translated, the model firstly encodes the sentence with a shared encoder. Then the shared decoder will choose a vocabulary to generate the translation. The decoder chooses the vocabulary based on the language token in the beginning of the input sentence. 

As we expected, the bidirectional translation model supports not only source-to-target, target-to-source translation, but also source-to-source and target-to-target self translation (the following section will describe how we train the model with such ability). With the ability of self translation, the shared encoder model can learn consistent distributions for different languages.

\subsubsection{Shared Encoder Model Training}
%\label{sec:sh_enc_training}
Similar to previous section, we introduce our shared encoder model training mechanism by a bidirectional model as an example.
We follows the state-of-the-art NMT training scheme. The major differences between our shared encoder model training scheme and the Transformer model ~\cite{vaswani2017attention} lie on the usage of bilingual training data, subword technique~\cite{sennrich2015neural}, source/ target side vocabularies, and training loss. Suppose we need to train a shared encoder model for two languages $L_1$ and $L_2$: 
\begin{itemize}
    \item Training data. We divide the training data of the bidirectional model into four portions: source-to-target ($D_{L_1\rightarrow L_2}$), target-to-source ($D_{L_2\rightarrow L_1}$), source-to-source ($D_{L_1\rightarrow L_1}$), and target-to-target ($D_{L_2\rightarrow L_2}$). For each portion, we add an extra language token in the beginning of each source-side sentence to distinguish which language the input sentence will be translated into. Specifically, we de-noised the $D_{L_1\rightarrow L_1}$ and $D_{L_2\rightarrow L_2}$ data to make sure the model will not "simply copy the input"~\cite{vincent2010stacked}.
    \item Subword. We train the encoder side byte pair encoding (BPE) on the mixing of $D_{L_1}$ and $D_{L_2}$. For the decoder-side, we train BPE model separately for each language. 
    \item Vocabularies. Following the scheme of subword training, we train one encoder-side vocabulary and two separated decoder-side vocabularies for a bidirectional translation model.
    \item Training loss. Equation~\ref{loss} shows the training loss of shared encoder model, which is a linear combination of losses of different target language.
    \begin{equation}
    \label{loss}
        loss_{MT}=\sum_{i=1}^k \lambda_i loss_{MT,i},
    \end{equation}
    where $\sum_{i=1}^k \lambda_i = 1$.
    $k$ is the number of language directions, which is 4 in a bidirectional translation case. For example, $loss_{MT,1}$, $loss_{MT,2}$, $loss_{MT,3}$, $loss_{MT,4}$ represent the loss of translation direction $L_1\rightarrow L_2$, $L_2\rightarrow L_1$, $L_1\rightarrow L_1$, and $L_2\rightarrow L_2$, respectively. Each loss can be calculated by a cross-entropy based loss function~\cite{cho2014learning,sutskever2014sequence}, which is widely-used in neural machine translation.  
\end{itemize}

\subsection{Semantic Textual Similarity Network}
We encode each sentence to the same length of hidden states as the input by using the pretrained sentence encoder in the previous Section.
\begin{equation}
    \textbf{h}_t = SharedEncoder(s_t) \nonumber
\end{equation}
where $h_1 = (h_{1,1}, ..., h_{n,1})$ and $h_2=(h_{1,2}, ..., h_{m,2})$. We further apply intra-sentence and inter-sentence attentions to obtain sentence representations.

\begin{figure}[!hbt]
    \centering
	\includegraphics[scale=0.4]{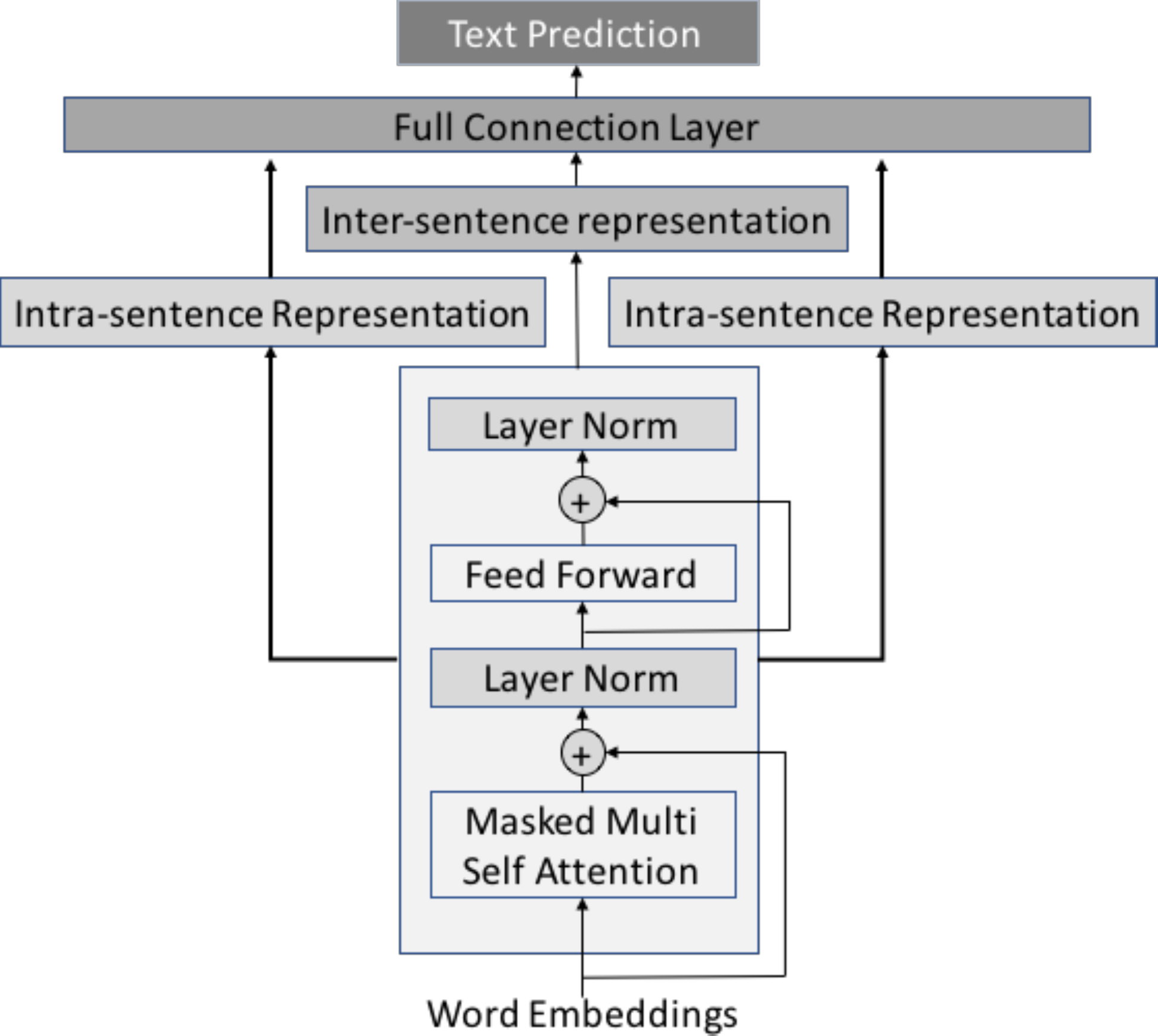}
	\caption{An overview of our shared encoder STS model architecture.}
	\label{fig:architecture}
\end{figure}

\subsubsection{Intra-sentence Attention}
We employ various aggregation methods, namely, max, mean and self-attention, and find self-attention works best. For self-attention aggregation, we adopt the attention mechanism from \cite{wang-EtAl:2017:Long2}:
\begin{align}
    s_{i,t} &=v^T{\tanh(W^ah_{i,t}+b^a)} \nonumber \\
    a_{i,t} &=\exp(s_{i,t})/\sum_{j=1}^n\exp(s_{j,t}) \nonumber \\
    v_{t}^{intra} &=\sum_{i=1}^na_{i,t}h_{i,t}
\end{align}

\subsubsection{Inter-sentence Attention}
Inter-sentence attention follows the approach described in \cite{DBLP:conf/coling/WangMI16}. We firstly calculate the semantic matching vectors for each sentence pair by soft aligning elements of one sentence to another:
\begin{align}
    e_{ij} &= F_{match}(h_{i,1})F_{match}(h_{j,2}) \nonumber \\
    \hat{h}_{i,1} &=\sum_{j=1}^m\frac{\exp(e_{ij})}{\sum_{j=1}^m\exp(e_{ij})}h_{j,2} \nonumber \\
    \hat{h}_{j,2} &=\sum_{i=1}^n\frac{\exp(e_{ij})}{\sum_{i=1}^n\exp(e_{ij})}h_{i,1} \nonumber
\end{align}
where $F_{match}$ function is a feed forward network.

After semantic matching phase, sentence vectors and its semantic matching vectors are compared and decomposed into similar components $h_{i,t}^+$ and dissimilar components $h_{i,t}^-$:
\begin{align}
    [h_{i,1}^+, h_{i,1}^-] &= F_{decomp}(h_{i,1}, \hat{h}_{i,1}) \nonumber \\
    [h_{i,2}^+, h_{i,2}^-] &= F_{decomp}(h_{i,2}, \hat{h}_{i,2}) \nonumber 
\end{align}
where $F_{decomp}$ is orthogonal decomposition to decompose $h_{i,t}$ into the parallel and perpendicular components with respect to $\hat{h}_{i,t}$.

Similar components and dissimilar components are further passed through $F_{comp}$ to obtain comparison representations and followed by an average pooling layer to aggregate all the features to obtain the inter-sentence representations:
\begin{align}
    v_{i,t}^{inter} &= F_{comp}(h_{i,t}^+, h_{i,t}^-) \nonumber \\
    v_{t}^{inter} &= \sum_{i=1}^{n}v_{i,t}^{inter}/n
\end{align}

\subsubsection{Representation Layer}
Given the intra and inter sentence representations, we concatenate the following terms as the
sentence pair representation $v_s$: absolute difference and element-wise multiplication between two intra-sentence representations; two inter-sentence representations:
\begin{equation}(|v_1^{intra} - v_2^{intra}|, v_1^{intra} \circ v_2^{intra}, v_1^{inter}, v_2^{inter})\end{equation}

\subsubsection{Output Layer}
The output layer exploits a fully connected neural network with two layers. The first layer uses 300 units with $relu$ activation function. The second layer has K units output combined with $softmax$ to produce probability distribution $\hat{p}$ on $K$ similarity labels. And the loss for STS task is computed as KL divergence between the predicted $\hat{p}$ and the true probability distribution $p$ which is verified in \cite{tai2015improved} to have better performance than squared error objective. And the predicted rating $\hat{y}$ is reconstructed from $\hat{p}$ by multiplying it with $r^T=[0,1,...,K]$ where $K=5$ for SemEval task.

%\begin{equations}
    \begin{align}
        & v_s' =relu(W^ov_s+b^o) \nonumber \\
        & \hat{p} = softmax(W^pv_s'+b^p) \nonumber \\
        & \hat{y} =r^T\hat{p}
    \end{align}
%\end{equations}

\subsection{Ensemble with Multilingual Sentence Representations}
For STS task in low-resource language $L_1$, one approach to augment data is to translate all the data set to another language especially to a rich-resource language $L_2$. Different models can then be trained with data sets in different languages and ensembled together to achieve better performance. This approach has been shown to be able to bring significant improvement on STS task in \cite{duma2017sef} where sentence representations are trained and calculated in different languages. And their average on consine similarity is used to measure the semantic similarity. However, this approach requires to call external translation service multiple times to translate sentences into different languages and thus incur large amount of extra response time in practice.

One beneficial brought by the pretrained shared encoder is that for each sentence, multiple output representations can be easily obtained without calling external translation service by simply prepending target language token to sentence. For example, with prepended token as language $L_1$, the encoder output will be in $L_1$ semantic space; with prepended token as $L_2$, the encoder output will be in $L_2$ semantic space.

\begin{equation}
    h_t^{L_i} = SharedEncoder([\langle L_i\rangle, s_t])
\end{equation}
Hence, besides the intra and inter sentence representations, we can further exploit this nature of shared encoder to obtain language-wise features. We use the representations in different language semantic spaces to calculate their prediction of probability distributions on semantic similarity level $\hat{p}_i$ and ensemble their predictions by linear combination as the final prediction:
\begin{equation}
    \hat{p} =\sum_{i=1}^n\beta_i\hat{p}^{L_i}
\end{equation}
where $\sum_{i=1}^n\beta_i=1$, $n$ is the number of languages that the shared encoder can support. And finally we calculate KL divergence between $p$ and $\hat{p}$ as similarity task loss $loss_{STS}$:
\begin{equation}
    loss_{STS} = KL(p||\hat{p})
\end{equation}

\begin{figure}[!ht]
	\centering
	\includegraphics[scale=0.4]{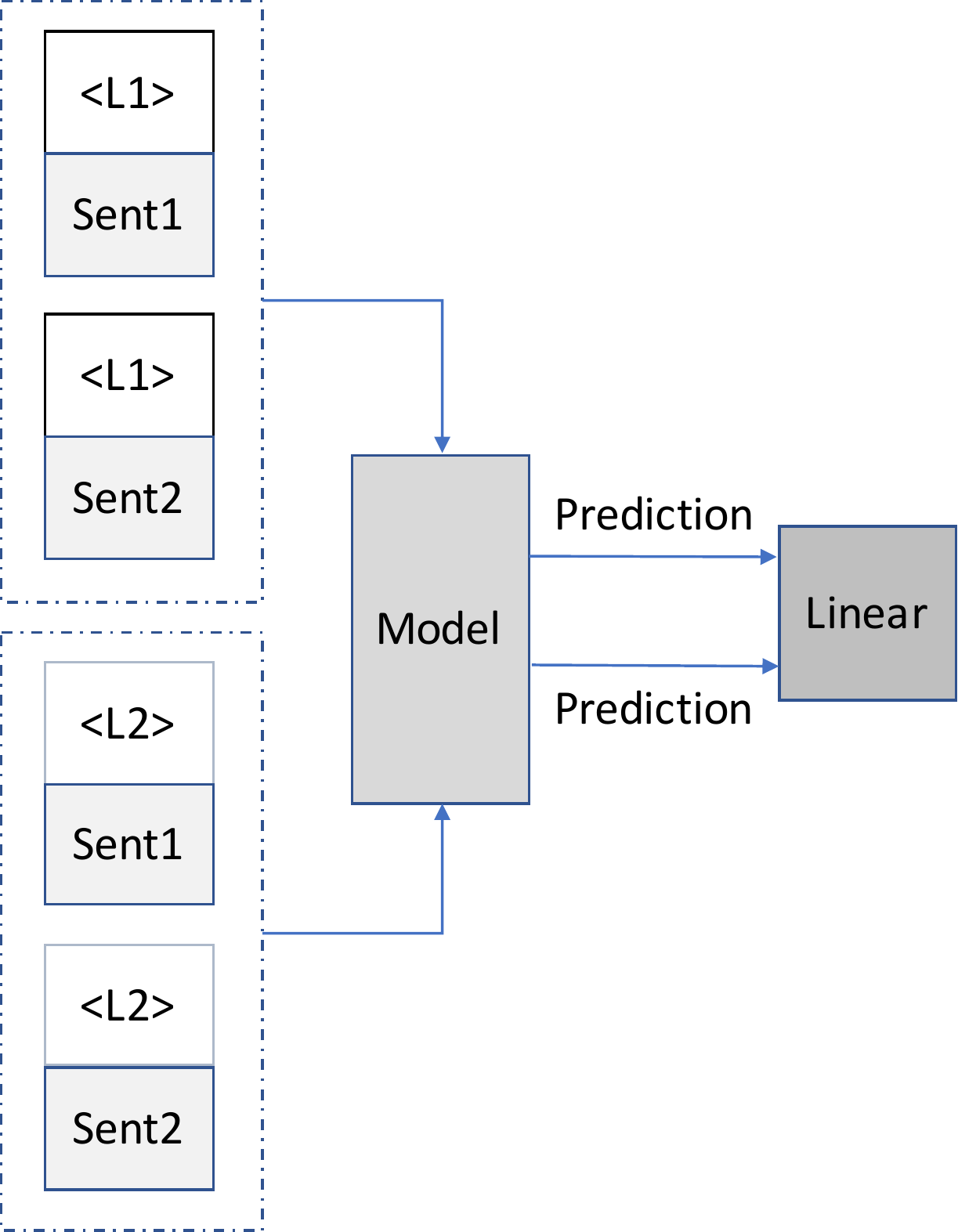}
	\caption{Ensemble of predictions from sentence representations in different language semantic spaces.}
	\label{fig:ConvEnc}
\end{figure}

\section{Experiments}
\subsection{Data Set}
We firstly verify our approach on public benchmark data set SemEval 2017 task 1 track-1(ar-ar) and track-3(es-es). And then it is also validated on an Indonesian and Thai data set of an industrial application. 

\subsubsection{SemEval Data Set}
SemEval-2017 task 1 requires to measure the relatedness of two sentences as score ranging from 0 for no meaning overlap to 5 for meaning equivalence. Table 1 shows the statistics of this data set. There are above 10,000 annotated data for English collected from the past SemEval STS task (2012-2015) while just around 1000 pairs for Arabic and Spanish which represents low-resource cases. Since SemEval provide no dev set for model evaluation and selection, we randomly select 20\% pairs from Spanish and Arabic training data as dev sets and exclude them from training.

All the training data is simply lower-cased, tokenized by white space and then byte pair encoded \cite{sennrich2015neural} to reduce the vocabulary size needed for translation pretraining. No hand-crafted feature is added.

\begin{table}  
\begin{tabular*}{8cm}{p{2.3cm}p{1.5cm}p{1.5cm}p{1.5cm}}
\hline  
    \textbf{Language Pair} &\textbf{Train} & \textbf{Dev} &\textbf{Test}\\  
\hline  
AR-AR&864&217&250\\
\hline
ES-ES&1244&311&250\\
\hline
EN-EN&12000&1592&250\\
\hline
\end{tabular*}  
\caption{Statistics of SemEval 2017 data set}
\end{table} 

\subsubsection{Industrial Data Set}
We also examine this approach on an e-commerce chatbot scenario that the target language Indonesian has about 4,000 labeled similarity pairs and Thai has about 10,000 labeled similarity pairs while English instance has already accumulated above 0.2 million. The data set was constructed based on online chatlog of a QA chatbot. If the answer of chatbot is correct, the user query and the knowledge title of the answer is labeled as similar. Otherwise, the label is dissimilar. We split the data set by date, saving 3 days of data as development set and 3 days of data as test set such that the data set can reflect the actual data distribution online.

The key difference between industrial data set and SemEval data set is that industrial data set contains large amount of abbreviations, spelling errors and grammar errors. Table 2 shows sentence from Indonesian data set and its translations by different translators. From the translation results, we can tell that translator introduce the errors like keeping the abbreviation untranslated or translating the misspelled word wrongly to other meaning.

\begin{table}
\begin{tabular*}{8cm}{p{1.5cm}p{5cm}}
\hline  
    &\textbf{Example sentences}\\  
\hline  
Original&Bgm cara sy memesa n bgm pembayran nya mks\\
Bing&BGM how sy n his pembayran bgm ordering mks\\
Google&How do I manage how to pay it?\\
Human&How can I order and how to pay it?\\
\hline
\end{tabular*}
\caption{Comparison of example Indonesian user query and its translations}
\end{table}  

\subsubsection{Machine Translation Data Set}
We used Paracrawl data~\footnote{\label{myfootnote}Provision of Web-Scale Parallel Corpora for Official European Languages, 
          \footnote{https://paracrawl.eu/}}, which contains about 16 million parallel sentence pairs for English $\leftrightarrow$ Spanish model training. OpenSubtitle 2018 portion of OPUS ~\cite{TIEDEMANN12.463} is used for English $\leftrightarrow$ Arabic model training, which contains about 31.9 million parallel sentence pairs. Both the test data consist of 1,000 randomly sampled bilingual sentence pairs from the corresponding training data. We used 50,000 BPE operations for source vocabularies and 30,000 for target vocabularies for both models. Top 50,000 and 30,000 tokens are kept for source and target vocabularies. 
\subsection{Parameter Setting and Evaluation Metrics}
Transformer sentence encoder uses the transformer-base setting from \cite{vaswani2017attention}: 512 hidden size, 512 embedding size, 2048 filter size, 8 heads for multihead attention, 6-layer encoder and 6-layer decoder. We train all the models with a batch size of 4096 tokens, and 0.0003 learning rate with Adam optimizer. We use case-insensitive BLEU-4 as our evaluation metrics for machine translation. 

When training semantic similarity task, the parameters of transformer encoder are fixed and only STS task specific parameters are trainable. All the feed-forward neural network is two layers with tanh as activation function. Adam optimizer is used with 0.0003 learning rate and batch size of 16. Early stop is used by observing the evaluation metrics on development set. We choose the model performed best on development set for later evaluation on test set.

The evaluation metrics for SemEval task is the pearson coefficient score $r$ between the gold rate $y$ and the predicted score $y'$.
\subsection{Results}
We include two Bag of Words models as baseline since BoW model is well known to perform strongly on semantic similarity task as they capture word identity information: (i) \textbf{one-hot embedding average}. Sentence representation is obtained by taking each dimension as whether an individual word appears in the sentence; (ii) \textbf{fasttext word embedding average}. Fasttext pretrained embedding \cite{bojanowski2017enriching} is used in this setting. Both methods use cosine value over two sentence representations to measure similarity.
\subsubsection{Machine Translation Results}
\label{sec:mt_result}
We firstly describe our results on machine translation in this section. Table 3 shows the MT evaluation on different language pairs. As we can see from the paper, all the models can get over 0.95 BLEU score on self-translation, which means our shared model can effectively learn the information of input sentence. Compared with single models (one specific translation direction), our shared models can achieve comparable or even higher BLEU score on bidirectional translation. 
% An interesting part is En-Tr model, where the shared model achieves much better results on bidirectional translation than the single model. As we described above, there are only about 200,000 bilingual sentence pairs for training En-Tr translation model. We believe that the self-denoised data and bidirectional data can provide more information in a low-resource translation scenario.

\begin{table}[h]
\label{table:mt_exp}
\centering
\begin{tabular}{p{1.5cm}|p{1cm}|p{1cm}|p{1cm}|p{1cm}}
\hline
Language 
Pairs         & Source & Target & BLEU-shared & BLEU-single \\ \hline
\multirow{4}{*}{EN-ES} & EN     & ES     & 0.4722      & 0.4744      \\ \cline{2-5} 
                       & ES     & EN     & 0.3592      & 0.3753      \\ \cline{2-5} 
                       & EN     & EN     & 0.9678      & NIL         \\ \cline{2-5} 
                       & ES     & ES     & 0.9884      & NIL         \\ \hline

\multirow{4}{*}{EN-AR} & EN     & AR     & 0.0978      & 0.1201      \\ \cline{2-5} 
                       & AR     & EN     & 0.3474      & 0.3415      \\ \cline{2-5} 
                       & EN     & EN     & 0.9890      & NIL         \\ \cline{2-5} 
                       & AR     & AR     & 0.9683      & NIL         \\ \hline
\end{tabular}
\caption{Machine translation results. For each language pair, there are 4 translation directions. For example, EN-ES represents the 4 translation directions related to English and Spanish: EN$\rightarrow$ES, ES$\rightarrow$EN, EN$\rightarrow$EN, and ES$\rightarrow$ES, respectively. BLEU-shared and BLEU-single represent the BLEU score of our shared translation model and single translation model baseline on each translation direction, respectively. As it is meaningless to train a self-translation single model, statistics about self-translation of single models are not shown in this table.}
\end{table}

\subsubsection{Comparison with Unsupervised Methods} 
From the experiment, BoW models give strong baselines for SemEval 2017 ar-ar, es-es and en-en test sets. One-hot embedding performs better than fasttext embedding on all three languages meaning that the lexical overlap is a quite strong feature for judging similarity on SemEval task. BoW baseline model also outperforms non-MT method of HCTI on Arabic and Spanish since for these two languages, there are very few training data about 1000 pairs. The full model with translation task pretrained encoder has significantly higher results than the two BoW baselines in all languages. However, among all the tasks, Arabic has the smallest improvement from 0.604 to 0.650 only. We argue that this may result from the discrepancy between English and Arabic is larger than English and Spanish which we can tell from the BLEU score of translation task is lower for En-Ar compared to En-Es.

\subsubsection{Comparison with Supervised Methods} 
In this work, we focus on methodology that does not require translating low-resource language to rich-resource language in inference time otherwise it will increases response time and largely depends on the translation quality of a third party system. However, from published results, we can see that MT based method achieve the highest scores on all tasks since it benefits from the large amount of training data in resource rich language which is English for SemEval task. Our model can achieve the same performance for Spanish as 0.825 which is quite close to 0.826 of MT method. However, for Arabic, the gap is still large. For HCTI non-MT method, it even cannot beat baseline BoW model for Arabic and Spanish test data which reaffirms that the training data of Arabic and Spanish is insufficient to train a supervised model from scratch. With the same insufficient training data of Spanish and Arabic, our approach shows significant improvement about 0.2 increasing on Pearson correlation score by incorporating more training data from English with MT pretrained encoder. From above comparison, we can conclude our approach can successfully transfer the knowledge in resource rich language to resource low language especially when two languages are closed to each other and can be easily translated to each other.

\begin{table}  
\begin{tabular*}{8cm}{p{2.3cm}p{1.5cm}p{1.5cm}p{1.5cm}}
\hline  
    &\textbf{AR-AR} & \textbf{ES-ES} &\textbf{EN-EN}\\  
\hline  
\multicolumn{3}{l}{\textbf{BoW Baseline}}\\
One-hot&0.604&0.711&0.727\\
FastText&0.549&0.686&0.559\\
\hline
\multicolumn{3}{l}{\textbf{MT Method}}\\
HCTI&0.713&0.826&0.811\\
\hline  
\multicolumn{3}{l}{\textbf{non-MT Method}}\\
HCTI&0.437&0.671&0.815\\
Our model&0.650&0.825&0.817\\

\hline
\end{tabular*}  
\caption{Pearson correlation coefficients comparison on STS 2017 task 1 ar-ar, es-es and en-en. HCTI model\cite{shao2017hcti} includes both word embeddings and hand-crafted features as its input to the model.}
\end{table}  

 \begin{table}  
\begin{tabular*}{8cm}{p{3cm}p{2cm}p{2cm}}
\hline  
    &\textbf{ID-ID} & \textbf{TH-TH}\\  
\hline  
\textbf{Baseline}&&\\
FastText&0.617&0.598\\
\hline  
\textbf{MT Method}&&\\
Decomposable Attention&0.507&0.528\\
\hline  
\textbf{Our Model}&&\\
w/o Pretrained&0.663&0.696\\
Pretrained&0.758&0.782\\
\hline
\end{tabular*}  
\caption{Result on Industrial Indonesian and Thai Data Set using AUC as   evaluation metrics. Decomposable attention refers to  \cite{parikh2016decomposable}}
\end{table}  

\subsubsection{Industrial Application Result}
 The result shows the limitation of MT based method which translates resource low language to resource rich language and predict. MT method performs well on SemEval task since the data set contains no abbreviation, spelling error and grammar error which make it easier to translate. However as illustrated in table 1, industrial data contains large amount of informal and incorrect words which largely affect the quality of translation. Table 5 shows MT method even performs much worse than the word average baseline. For our approach, with MT pretrained, the AUC score can be improved from 0.617 to 0.758 for Indonesian, 0.598 to 0.782 for Thai. In both industrial data set we have consistently improvement.

\subsubsection{Impact of Number of Finetuned Layers}
For above experiments, when training semantic similarity task, parameters of pretrained sentence encoder are fixed. We also observed the impact that if we have different number of layers of sentence encoder trainable. We unfreeze starting from the last layer, last 2 layers and finally to all 6 layers since last layer contains least
general knowledge \cite{yosinski2014transferable}. The result shows that the performance fluctuates and decrease a lot if all layers are trainable. Unfreezing the last several layers can improve the performance for some settings but it does not have a consistent pattern across all languages. This observation is not consistent with the one in \cite{radford2018improving} that supervised finetuning the pretrained language model in target task can bring extra benefits as the number of layers increases. We argue that it is because STS task has relative small data set about 10k and it is more prone to overfit.

\begin{figure}[!ht]
	\centering
	\includegraphics[scale=0.4]{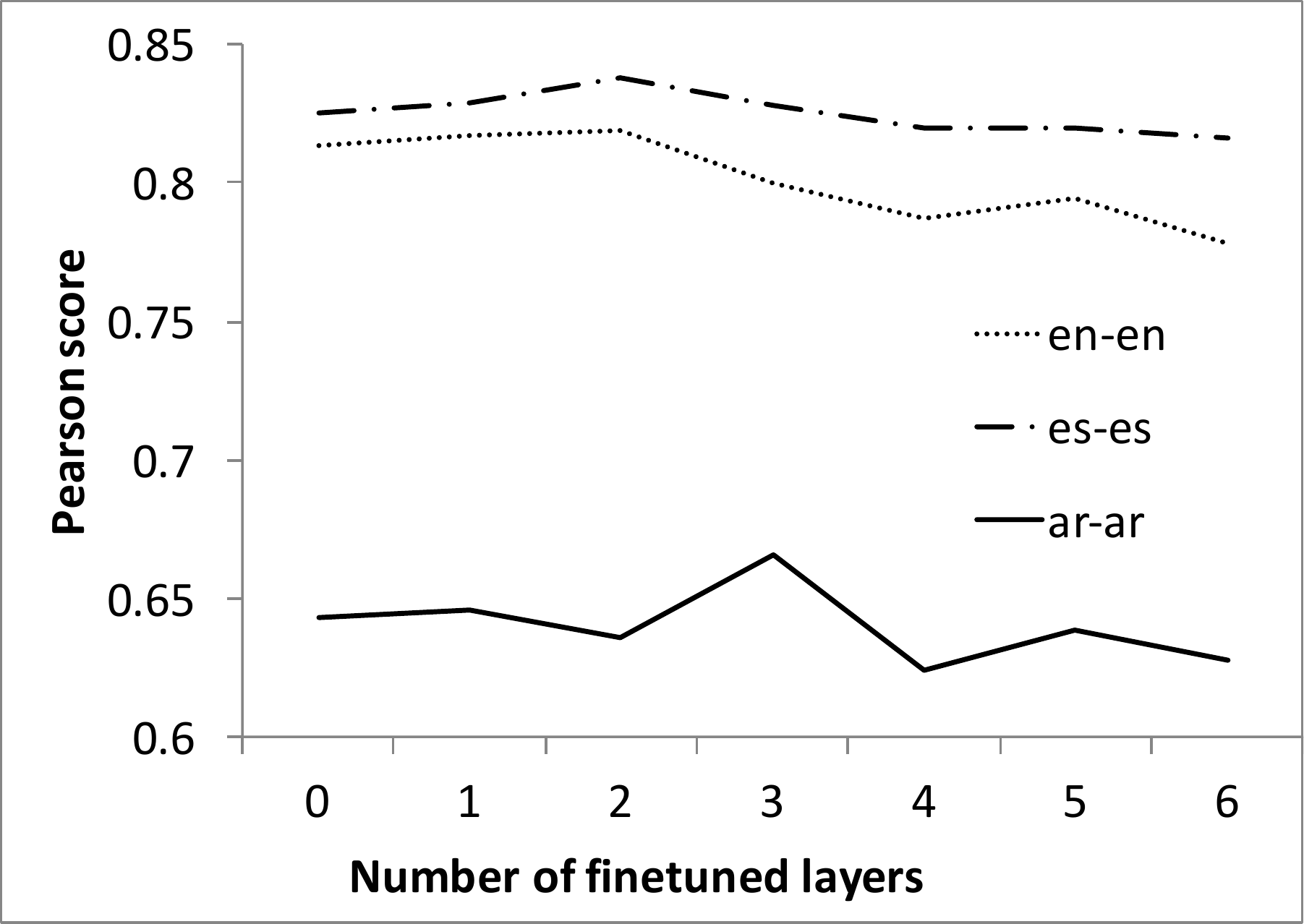}
	\caption{Effects of number of finetuned transformer encoder layers. The variable number means that we unfreeze the layers starting from no layer, the last layer, and finally to all layers.}
	\label{fig:finetuned}
\end{figure}

\subsubsection{Ablation Study}
We perform an ablation to study the contribution of different tasks of our methodology. We train multiple models with missing different training data and part of the model: rich-resource training data, low-resource training data and multilingual sentence representations.

Without pretraining the sentence encoder, we can see that for Spanish training data only setting, the Pearson score is very low compared to English training data only setting since the small number of Spanish training data is inadequate to train a supervised model. And the performance does not improve when combining English and Spanish training data together for model without pretraining. For each setting, after pretraining, the performance has significant improvement especially for Spanish training data only from 0.188 to 0.727 which also exceeds non-MT HCTI method. This confirms that MT pretraining task can help to improve the performance when supervised training data is insufficient. And for the pretrained model, adding English data and multilingual representations also further contribute to improve the performance.

\begin{table}  
\begin{tabular*}{8cm}{ccc}
\hline  
    &\textbf{w/o Pretrained} & \textbf{Pretrained}\\  
\hline  
En only&0.668&0.766\\
Es only&0.188&0.727\\
En + Es&0.661&0.796\\
+multilingual repr&0.6738&0.825\\

\hline
\end{tabular*}  
\caption{Ablation study on SemEval 2017 task es-es}
\end{table}

\section{Related Works}
%\label{sec:relate}

%We briefly describe the related works in several areas in this section.

%\subsubsection{Semantic Textual Similarity: } 
In this paper, we focus on the problem of semantic textual similarity, which is widely applied in many scenarios. However, most researches work on the supervised settings of monolingual language. \cite{DBLP:conf/emnlp/HeGL15} adopts CNN to capture features at multiple granularities for comparing sentence representations by using multiple similarity metrics. While in \cite{DBLP:conf/coling/WangMI16}, not only the similar parts of two input sentences but also the dissimilar parts are taken into account by decomposing and composing lexical semantics over sentences. Although outperforming many traditional methods, these prior works rarely consider the impact of the other sentence when deriving the sentence representation. Until in \cite{DBLP:journals/tacl/YinSXZ16}, an attention-based model is proposed to use the content of one sentence to guide the representation of the other, in which an attention feature matrix is learned to influence the convolution filters. Different to the previous methods, \cite{DBLP:conf/aaai/PangLGXWC16} takes into account the rich interaction structures in the text matching process since the interaction structures are compositional hierarchies in which higher level signals are obtained by composing low level signals. All these methods are based on the convolutional neural network and trained with a large scale labeled data. However, only a few labeled data is provided in our scenario and we have to solve the low-resource problem.

Universal sentence encoder is also a popular research topic in recent years. Most works are usually based on a multi-task learning framework. In \cite{DBLP:journals/corr/abs-1803-11175}, two variants of encoding models, Transformer \cite{vaswani2017attention} and Deep Averaging Network \cite{iyyer2015deep}, allow for the trade-offs between accuracy and efficiency of diverse tasks, such as sentiment analysis and natural language inference. \cite{DBLP:journals/corr/abs-1804-00079} exploits the effectiveness of inductive biases in the context of a simple one-to-many multi-task learning framework. In their work, a single recurrent sentence encoder is shared across multiple tasks, which are skip-thoughts, machine translation, natural language inference and constituency parsing. As shown in the above works, sentence-based approaches are universal to different tasks while not to different languages. 

In recent years, more and more studies \cite{Peters:2018,howard2018universal,radford2018improving} show that pretraining a universal encoder with large-scale unlabeled data and then finetuning on a task-specific network with supervision are effective for most tasks. All these studies leverage universal language model as the unsupervised pretraining model to capture more linguistic information which are useful to many downstream tasks. Our method in this paper is closest to these frameworks. However there are still two differences. We aim to build up a multilingual sentence encoder by taking a shared encoder translation model as the pretraining model. Multiple sentence representations can be generated by our framework and naturally ensembled to fit the target task. To the best of our knowledge, this paper is the first to study the multilingual sentence encoder for semantic textual similarity.
\section{Conclusion}
In this paper we propose a solution to improve the multilingual semantic textual similarity in low-resource languages by using a shared sentence encoder.  The shared encoder is pretrained via a bi-directional and self denoising task to enable its multilinguality.  By using this shared encoder, we can obtain various sentence representations for a sentence in different language-specific semantic space, and utilize them in an ensemble model for better performance in similarity evaluation.  Experimental results show that our model consistently beats state-of-the-art non-MT approaches, and even reach the same performance of MT-based methods in Spanish task. It is noteworthy that our framework is a generic approach to construct multilingual sentence representation requiring no language specific prepossessing and hand-crafted features. 

%Although in this work we are focusing on STS in Spanish and Arabic, our method shall also work well in other languages and other sentence representation tasks.

%This paper proposes a multilingual sentence encoder for Semantic Textual Similarity (STS). Although supervised approaches are well studied, little attention is paid to languages with limited resource. In this work, we tackle this problem with a multilingual framework to utilize the rich-resource in a source language. 
%In order to alleviate language discrepancy, we employ a shared encoder on both languages. To pretrain this encoder, we conduct bi-directional and self denoising task to achieve multilingual encoder. With the nature of a shared encoder, sentence representations in different language semantic space can be obtained and ensembled together for better performance.

\section*{Acknowledgments}

\bibliographystyle{aaai}
\bibliography{aaai19}

\end{document}